\documentclass[runningheads]{llncs}
\usepackage[T1]{fontenc}
\usepackage{graphicx}
\usepackage{tabularx}
\usepackage{amsmath}
\usepackage{tabularray}
\usepackage{multirow}
\usepackage{appendix}
\usepackage[table,xcdraw]{xcolor}
\usepackage{color}
\usepackage{hyperref}
\hypersetup{
    colorlinks=true,linkcolor=blue,citecolor=blue,urlcolor=blue
}

\usepackage{tipauni}
\usepackage{fontspec}
\newfontface{\bn}{kalpurush.ttf}
\newfontface{\ipa}{DoulosSIL-Regular.ttf}

\begin{document}
\title{Transcribing Bengali Text with Regional Dialects to IPA using District Guided Tokens}
\titlerunning{Transcribing Bengali Text with Regional Dialects to IPA using District Guided Tokens}

\author{S M Jishanul Islam\inst{1} \and
Sadia Ahmmed\inst{2} \and
Sahid Hossain Mustakim\inst{1}
}
\authorrunning{Islam et al.}

\institute{\textsuperscript{1}United International University \\
\textsuperscript{2}University of British Columbia \\
Correspondence: \email{\{sislam201024\}@bscse.uiu.ac.bd}\\
}

\maketitle              
\begin{abstract}
Accurate transcription of Bengali text to the International Phonetic Alphabet (IPA) is challenging due to the language's complex phonology and context-dependent sound changes. This challenge is aggravated in regional Bengali districts due to the unavailability of standardized spelling conventions, the presence of regionally popular local and foreign words, and phonological diversity across different regions. This study presents the first approach to tackle the never-before-attempted sequence-to-sequence problem of transcribing regional Bengali text to IPA by introducing District Guided Tokens (DGT). Our key idea is to provide the model with explicit information about the regional district or ``district" to help understand the unique phonetic pattern associated with each district of the input text before generating the IPA transcription. We utilize a new dataset consisting of text-ipa pairs, spanning six districts of Bangladesh. Our experimental results after fine-tuning various transformer models demonstrate the effectiveness of DGT, with the token-free ByT5 model achieving superior performance over word-based models like mT5, and umT5. The three models with DGT outperform the others, with the byt5-DGT model attaining the best WER and CER rates of 1.20\% and 0.42\%, respectively. To accelerate the progress of regional Bengali district applications, we open-source our family of models to the research community.

\end{abstract}
\section{Introduction}

The International Phonetic Alphabet (IPA) provides a standardized representation of the sounds in spoken languages. Accurate text transcription to IPA is crucial for various applications like speech synthesis and linguistic studies \cite{mukherjeebengali}, \cite{sen2022bangla}. However, this task poses significant challenges for languages with complex and diverse linguistic variations (such as morphological, phonological, and syntactic) present in the districts of Bangla.

Bengali, the national language of Bangladesh which is native to the Bengal region of South Asia, exhibits several intricacies in its sound system. The language has an inherent consonant mutation phenomenon through Gemination, where sounds change at word boundaries based on the surrounding context \cite{dave2021neural}. Furthermore, certain Bengali characters can represent multiple sounds depending on the region or district. This regional influence on pronunciation is particularly evident across the diverse districts of Bangladesh. Alongside, some words heavily differ in spelling and pronunciation from standard Bengali in the case of almost every regional district. For example, the pronunciation of the word \bn{আমার} (my) in standard Bengali changes into \bn{আঁর} (\ipa{ɐ̃ɾ}) in Chittagong, \bn{মোর} (\ipa{moɾ}) in Rangpur, \bn{আমগো} (\ipa{ɐmgō}) in Narshindi, \bn{আর} (\ipa{ɐɾ}) in Narail, \bn{আর} (\ipa{ɐɾ}) in Tangail and \bn{আমার} (\ipa{ɐmɐɾ}) in Kishorganj. This distinction in phonetics across regional districts makes its transcription to the International Phonetic Alphabet (IPA) extremely perplexing and often completely unaddressed.

Text-to-IPA transcription is an important task with various existing approaches. Machine learning models for direct speech-to-IPA transcription aim to provide a universal and language-agnostic solution \cite{taguchi2023universal}. Recently, neural sequence-to-sequence models (Seq2seq) \cite{sutskever2014sequence} have shown promising results by learning the mapping between text and IPA representations from data. However, their performance is hindered by the presence of out-of-vocabulary (OOV) words and their inability to effectively capture the unique sound patterns and pronunciation differences across different regions. In this study, we propose a technique that uses District Guided Tokens (DGTs) to enhance the performance of sequence-to-sequence models for Bengali text-to-IPA transcription. We utilize the large-scale dataset proposed by \cite{fatema2024ipa} that contains Bengali regional text-IPA pairs. The core idea is to provide the model with explicit information about the input text's regional district before generating the IPA transcription. This is achieved by prepending a district token in the name of the district of origin to the input sequence, effectively guiding the model to capture the unique phonetic patterns associated with each district. We apply the DGT technique to fine-tune various encoder-decoder models on this dataset. The ByT5 \cite{xue2022ByT5} approach is particularly well-suited for this task, as it mitigates the issue of OOV words by operating directly on the byte sequences of the input text.

Through extensive experiments, we demonstrate the effectiveness of the DGT technique in improving the performance of seq2seq models for the Bengali text-to-IPA transcription task. By utilizing DGT in all three models—Byt5-small, umT5-base, and mT5-base—we achieve considerably improved outcomes, with a WER and CER rate of 1.20\% and 0.42\%, 2.39\% and 1.01\% and 27.79\% and 5.64\% respectively.  These results reflect the incorporation of regional district information into natural language processing systems for languages with diverse linguistic variations. Our contributions are as follows:
\begin{itemize}
    \item We propose a novel method called the District Guided Tokens (DGT) technique to improve the performance of sequence-to-sequence models for Bengali text-to-IPA transcription by explicitly incorporating regional district information. 
    \item We apply the DGT technique to various encoder-decoder models as a plug-in-and-play framework.
    \item We perform a detailed analysis of the models' performance against non-DGT methods and their performance for the districts of each regional district.
\end{itemize}

We release the checkpoints for all our models through our HuggingFace collections\footnote{\url{https://huggingface.co/collections/teamapocalypseml/bengali-regional-text-to-ipa-models-65eff2c76e38bf2ff9656442}}, as well as the training and inference notebooks through our GitHub repository\footnote{\url{https://github.com/S-M-J-I/bhashammo-dgt-regional-bengail-2-ipa}}.

The remainder of this paper is organized as follows. In Section~\ref{sec2}, we discuss the studies closely related to our field of work. Next, in Section~\ref{sec3}, we explore the dataset, discuss the pre-processing techniques, and declare the reasoning behind some key decisions. In Section~\ref{sec4}, we explain the methodology of our DGT technique in detail and the various transformer models used to model this task. We observe and report the performance of our technique in detail and validate its performance in Section~\ref{sec5}. We discuss the limitations and the solutions for mitigating them in Section \ref{sec6}. Finally, we present our conclusions and discuss the impact of this work in Section~\ref{sec7}.

\section{Related Works\label{sec2}}

Recently, some research efforts have been dedicated to addressing the challenge of translating regional districts of various languages worldwide. However, to date, very limited research has been conducted on the IPA transcription of the standard Bengali language, and there is still a notable void in the exploration of transcribing regional districts of Bengali.

Achieving mutual intelligibility between regional districts and a common language is a significant hurdle for low-resource languages like Bengali. To fill this gap, Faria et al. \cite{faria2023vashantor} introduce a large-scale dataset named "Vashantor" which translates five major regional districts of Bengali into standard Bengali. They employ SOTA translation models (mT5, BanglaT5 \cite{bhattacharjee-etal-2023-banglanlg}) and evaluate their performance using metrics like Character Error Rate (CER), Word Error Rate (WER), BLEU score, and METEOR score. Additionally, they utilize region detection models like mBERT and Bangla-bert-base to identify the specific regions associated with the input text. Their results demonstrate the effectiveness of these models, with mBERT achieving an accuracy of 84.36\% and Bangla-bert-base achieving an accuracy of 85.86\%. 

In recent years, efforts have been made to establish a consistent IPA standard for the Bengali language. Fatema et al. \cite{fatema2024ipa} presents a comprehensive study of the IPA standard for Bengali, discussing the existing challenges in representing Bengali sounds using IPA. They propose a consistent IPA transcription framework, addressing issues such as vowel representation, diphthongs, consonants, and diacritics. Additionally, the authors introduce a novel dataset of 150,000 Bengali sentences with their corresponding IPA transcriptions and provide benchmarks using a deep learning model for the task of Bengali-to-IPA transcription. Their work aims to facilitate linguistic analysis, natural language processing (NLP) resource creation, and downstream technology development for the Bengali language.

Recent advancements in Audio Speech Recognition (ASR) have enabled the development of universal models for transcribing speech into the IPA, a crucial step in language documentation and preservation efforts. Taguchi et al. \cite{taguchi2023} introduce a state-of-the-art model that outperforms existing speech-to-IPA models. Based on wav2vec 2.0 and fine-tuned semi-automatic IPA transcriptions from seven languages, their model outperformed existing speech-to-IPA models trained on larger datasets. Their research highlights the importance of high-quality phonetic transcriptions and linguistic diversity in training data. Additionally, their model approached human-level accuracy in transcribing unseen languages, demonstrating its potential for documenting understudied and endangered languages.

\begin{figure}[t]
    \centering
    \includegraphics[width=0.85\textwidth]{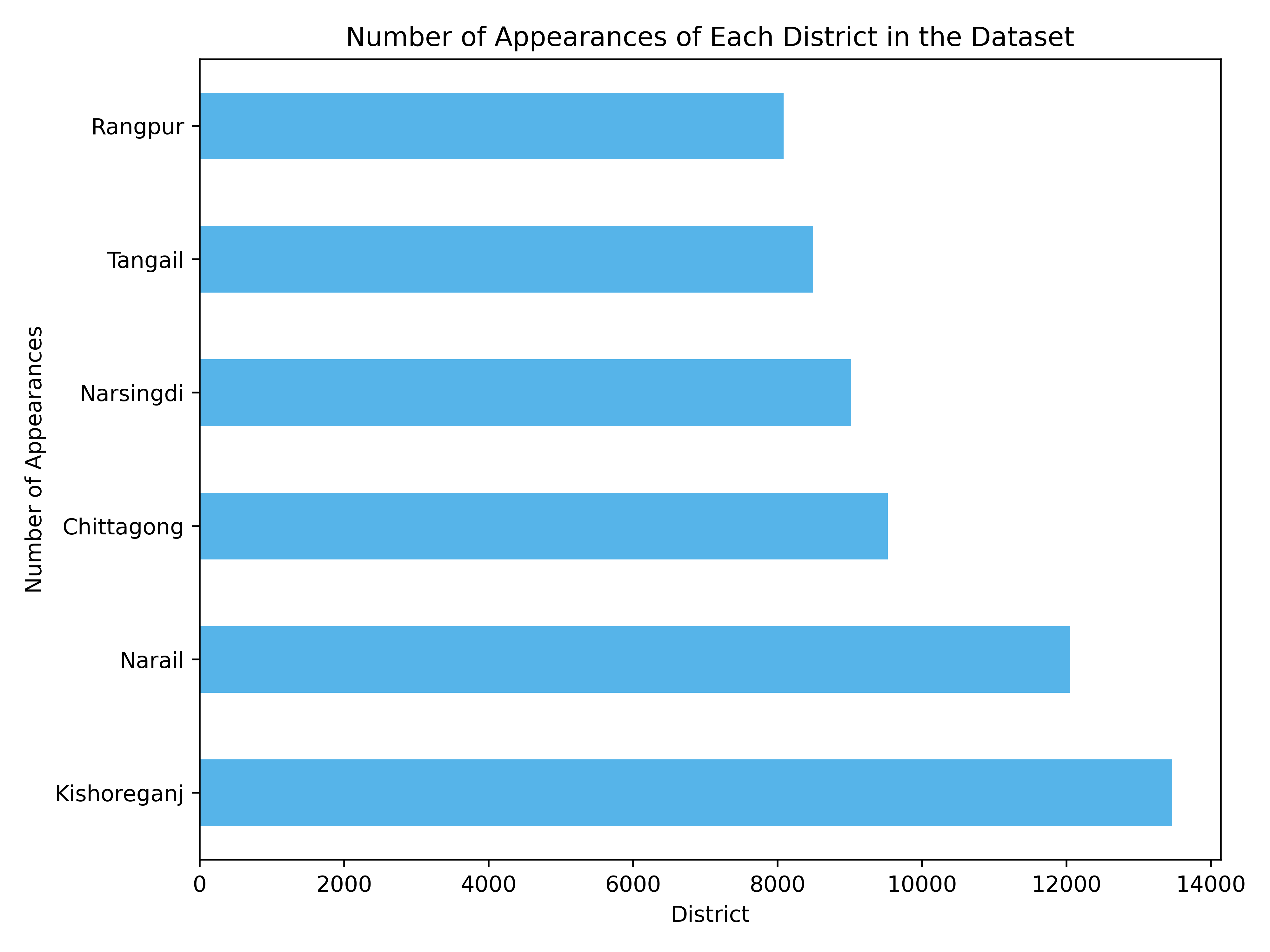}
    \caption{The frequency of districts appearing in the dataset.}
    \label{fig:data_districts}
    \vspace*{-\baselineskip}
\end{figure}

\begin{figure}[t]
    \centering
    \includegraphics[width=\textwidth]{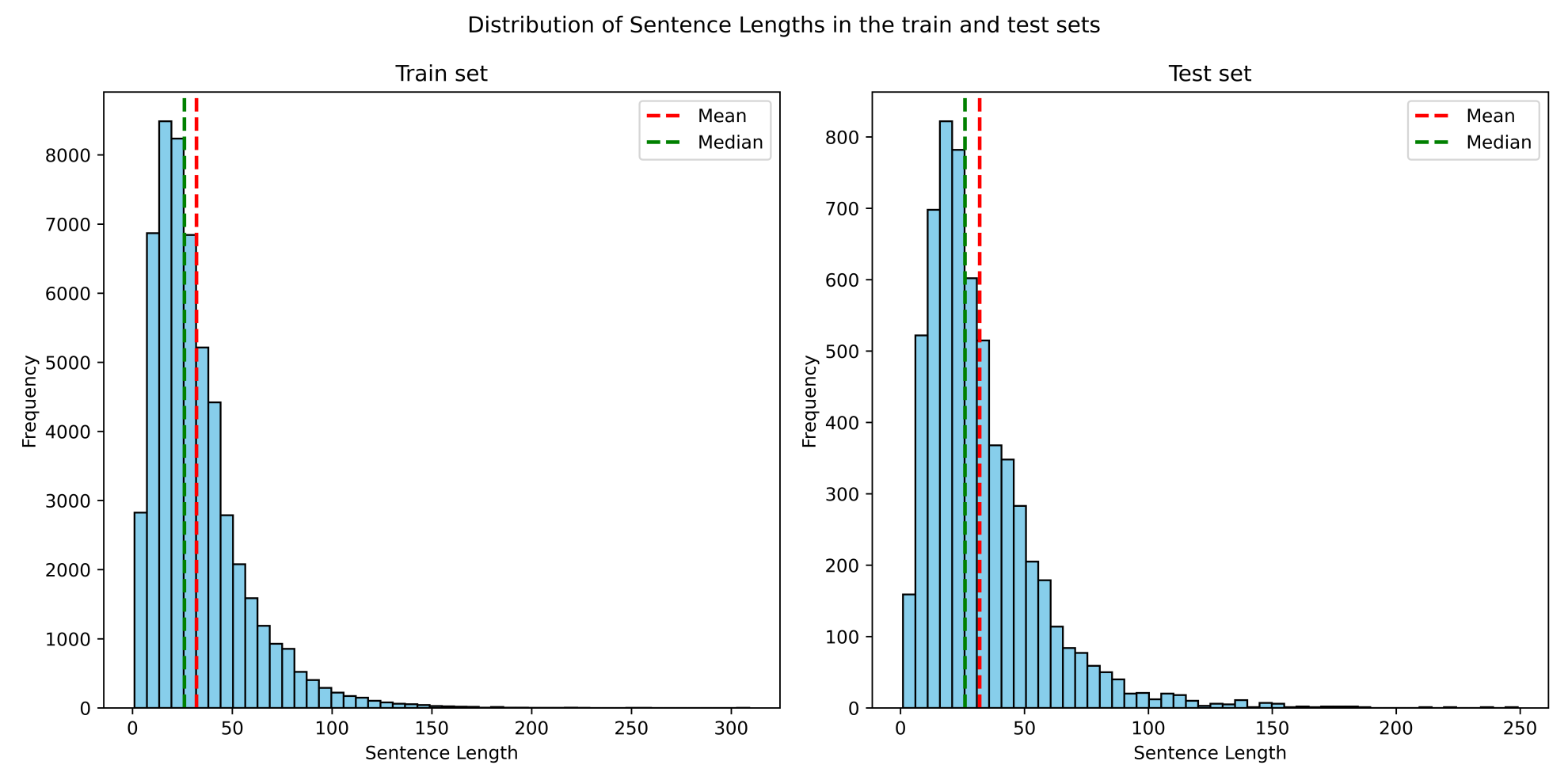}
    \caption{The number of distributions for the train and test set splits}
    \label{fig:data_trainvtest}
    \vspace*{-\baselineskip}
\end{figure}

\section{Dataset\label{sec3}}

In this section, we describe the dataset used in our study. The data comes from a study by Fatema et al. \cite{fatema2024ipa}. This dataset is a collection of Bengali text samples with regional districts and their corresponding IPA representations. The dataset covers district text-IPA pair samples from six districts in Bangladesh. We explore the dataset's features and the pre-processing methods required to make the input text clearer to the model. We divide the data into three sets for training, validating, and testing our model. We use 80\% of data for the train set, 10\% for the validation set, and the remaining 10\% for the test set.

\vspace*{-\baselineskip}

\subsection{Exploratory data analysis}

Before using the dataset examples for our study, we perform a brief exploratory data analysis to find its characteristics and set certain criteria for our methodology. This analysis informs us to make key decisions, such as setting the maximum generation length and opting for byte-based rather than word-based models.

We observe from Figure~\ref{fig:data_districts} that the district Kishoreganj is most represented while the district Rangpur is least represented. The figure clearly shows that the districts in the dataset are evenly distributed, and the performance of each district is not hampered due to one appearing less than the other. We validate this claim in Section \ref{sec5}.

Our analysis of the datasets' text lengths, shown by Figure \ref{fig:data_trainvtest}, reveals an average sentence length of approximately 32.08, with a median of 26. Most sentences fall below 100 words. Notably, the dataset holds a maximum input sentence length of 309 and a maximum output IPA length of 367. These lengths originate from distinct examples and are illustrated in Appendix \ref{appendix:longestsen} and \ref{appendix:longestipa} respectively. To establish a suitable generation length for model input during transcription, we set the models' maximum generation limit at 512 words.

Our analysis of this dataset also revealed a significantly high number of out-of-vocabulary (OOV) words. The training data vocabulary comprised 44,998 unique words, while the test set contained only 11,213. This means that 2,888 words (around 25.76\%) are not encountered during training. Therefore, we opted for a byte-based model for the main architecture, prioritizing character-based processing over word-based approaches to mitigate this problem.

\subsection{Pre-processing}

The correctness of our model's IPA generation is only as good as the quality of the input text data. To address this, we employ pre-processing techniques that enhance the clarity of the transcription produced by our models. This includes fixing English numerals in Bengali text and transforming multi-line sentences into one. We found examples that contained English numerals in the input text rather than the usual Bengali numerals. To show the model that these should be Bengali numerals rather than English, we manually annotate the examples to replace the English numerals with Bengali numerals. We have also found examples that have input sentences spanning multiple lines. This means that when the input sequence is processed as a string, it results in the appearance of newline characters. We mitigate this by replacing the newline characters with blank strings to create a single-sentence representation.

\section{Methodology\label{sec4}}

\begin{figure}[t]
    \centering
    \includegraphics[width=\textwidth]{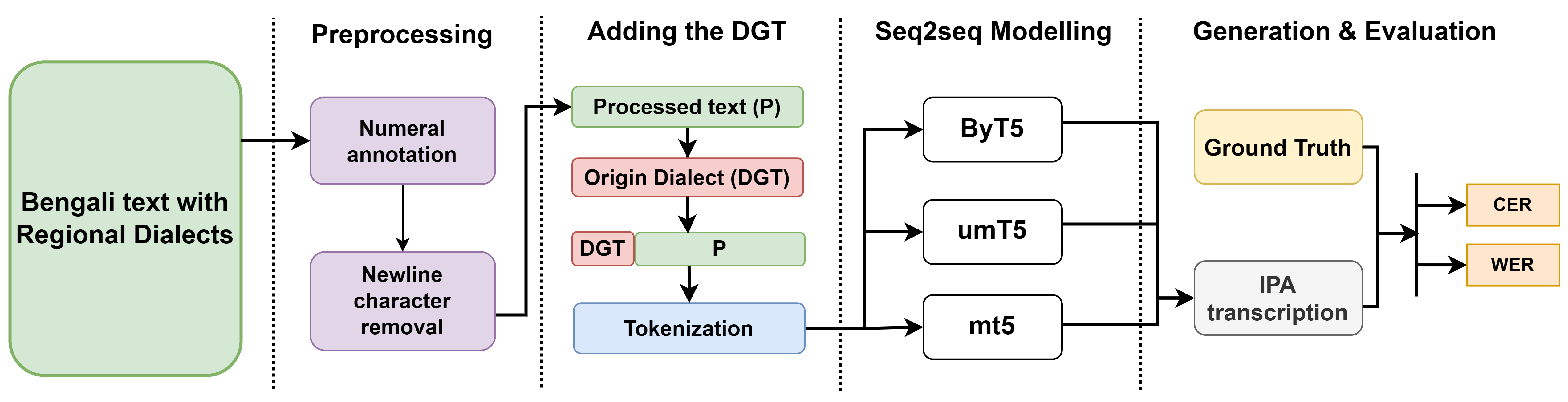}
    \caption{The end-to-end workflow for our methodology}
    \label{fig:pipeline}
    \vspace*{5pt}
\end{figure}

In this section, we explain our proposed methodology in detail. First, we explain the District Guided Tokens and then discuss the transformer models utilized in our implementation. An end-to-end workflow of our methodology is illustrated in Figure \ref{fig:pipeline}.

\subsection{District Guided Tokens}

Regional Bengali districts are often different from each other in terms of pronunciation. Even with identical written text, the desired pronunciation can vary depending on the region. This ambiguity can confuse models trying to generate accurate IPAs for different districts. Thus, we draw our ideas from the \verb|[SOS]| token and \verb|[CLS]| tokens from BERT \cite{devlin2018bert} and other transformer-based models, and propose new special tokens, known as District Guided Tokens, to carry the district information for a given input text so the model can follow along its district.

The District Guided Tokens (DGTs) help the models understand regional district variations in Bengali text. These special tokens, represented in the form of \verb|<district>|, are pre-pended at the beginning of a sentence. These tokens act like labels for the district and the origin district and are encountered by the model at the start of the generation process. When the model encounters this token, it acknowledges the district of origin and starts generating the IPAs from the input text sequence using its pronunciation rules, hence the term "Guided Tokens". The input to the model is transformed to the format: \verb|<district>| \verb|<Bengali text>|.

\begin{figure}[t]
    \centering
    \includegraphics[width=\textwidth]{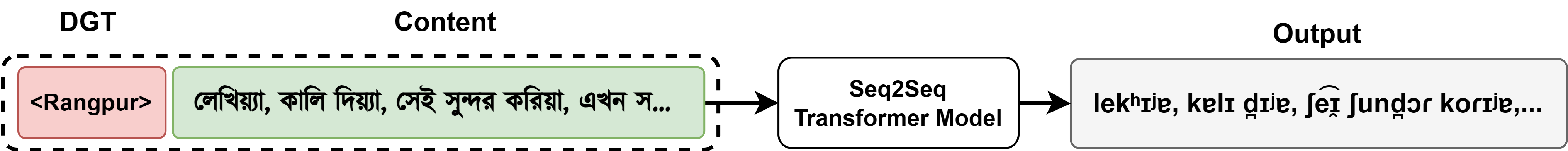}
    \caption{The illustration of the District Guided Tokens method}
    \label{fig:dgtdemo}
\end{figure}

During the tokenization process, the default tokenizers would treat these tokens as a series of encodings, rather than a single encoding. Therefore, we represent these tokens as "words" to the tokenizer's vocabulary. This would transform the DGT into a single encoding. Introducing these tokens as single words would also mean that the model needs to understand that they are single words. Thus, we increase the embedding size of the transformer model to adapt it to the new tokens so it can recognize their indexes during embedding. Since the dataset contains six districts of Bangladesh, it would result in the creation of six DGTs which are as follows: \verb|<Chittagong>|, \verb|<Narail>|, \verb|<Narsingdi>|, \verb|<Rangpur>|, \verb|<Tangail>|, and \verb|<Kishoreganj>|. The DGT methodology is illustrated in Figure \ref{fig:dgtdemo}.

\begin{figure}[t]
    \centering
    \includegraphics[width=\textwidth]{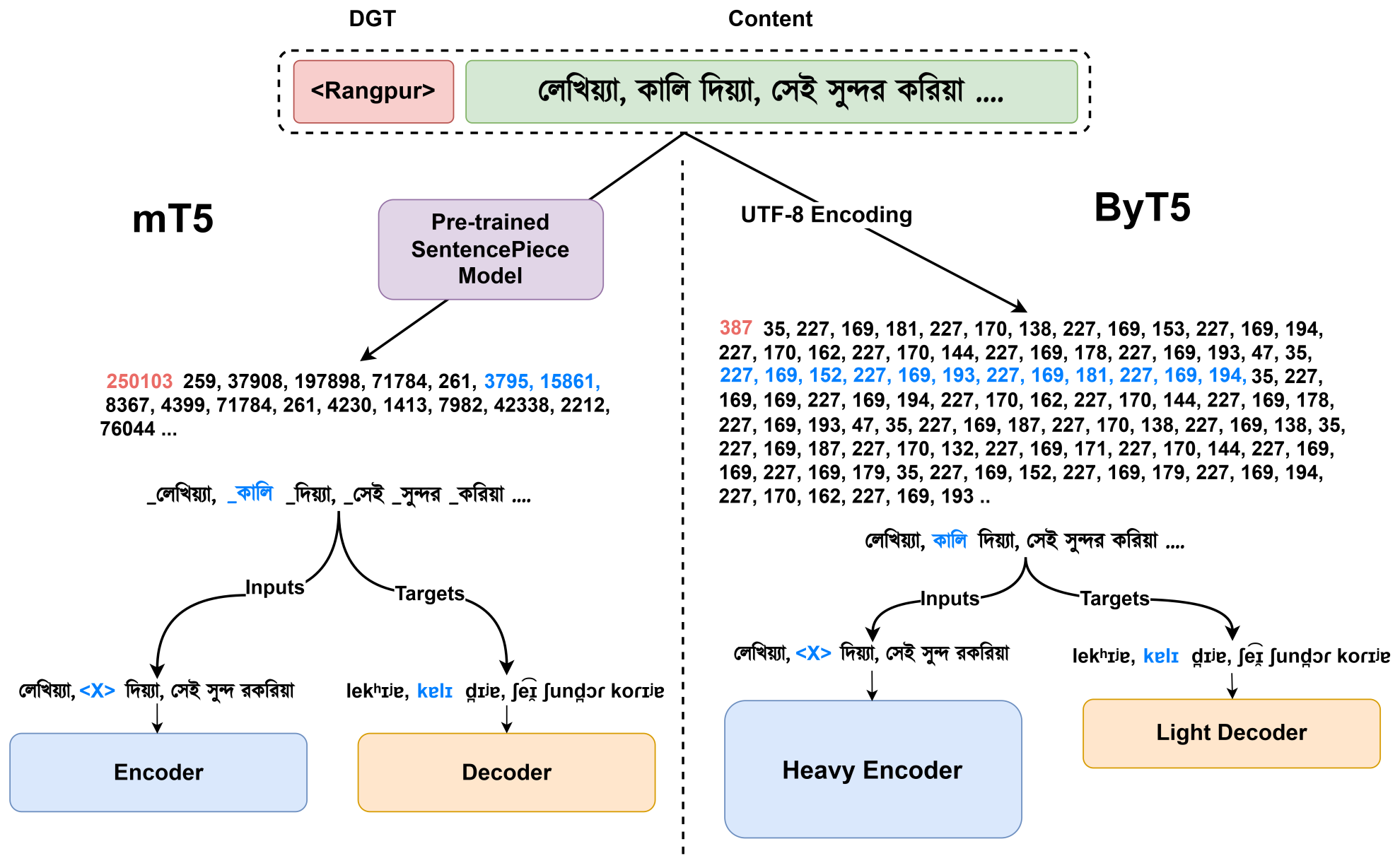}
    \caption{The mt5 \cite{xue-etal-2021-mt5} and ByT5 \cite{xue2022ByT5} architectures illustrated for this task. The encodings in red represent the unique indexes of the DGT for those specific models}
    \label{fig:models}
    \vspace*{-\baselineskip}
\end{figure}

\subsection{Sequence-to-sequence Transformer models}


The input is a Bengali text with regional districts, and the output is the text transcription as IPAs. Therefore, we formulate the task as a \textbf{sequence-to-sequence (seq2seq) text-to-text}. We choose three state-of-the-art transformer-based models with encoder-decoder architectures to model the formulated task. Since the Bengali language and phonetic alphabets are present in this task, we choose the multilingual version of these transformers. These transformers include the mT5, umT5, and ByT5 are illustrated by Figure \ref{fig:models}.

\vspace*{-\baselineskip}

\subsubsection{Massively Multilingual T5 (mT5)}

\cite{xue-etal-2021-mt5} is a multilingual version of T5 \cite{roberts2020knowledge} that underwent pre-training using a multilingual variant of the C4 dataset named mC4 \cite{2019t5} covering text in 101 different languages that is derived from Common Crawl web scrape. The mT5 model is built around the "T5.1.1" recipe, which enhances the T5 model by incorporating GeGLU non-linearities \cite{shazeer2020glu} and pre-training solely on unlabeled data without employing dropout. The vocabulary has increased to 250,000 wordpieces, and SentencePiece models trained using pre-training language sample rates are utilized. To handle languages with large character sets, it uses 0.99999 character coverage and SentencePiece's "byte-fallback" functionality to ensure string uniqueness. The model has undergone pre-training for 1 million steps using batches of 1024 input sequences, each with a length of 1024. This amounts to about 1 trillion input tokens in total. It is available in 5 different versions, small, base, large xl and XXL.

\vspace*{-8pt}

\subsubsection{UniMax mT5 (umT5)}

\cite{chung2022unimax} is an effective two-pronged method of sampling that achieves a more balanced representation of major languages and restricts overfitting on less common languages by limiting repeats in each corpus. Moreover, UniMax effectively addresses the model scalability issue by explicitly managing data repetition in any language. This approach directly prevents overfitting in low-resource languages without prioritizing higher-resource languages. UmT5 is derived from mT5 employs UniMax and incorporates a unique relative positioning bias that is calculated individually for each layer. It is pre-trained on the updated version of the mC4 corpus, covering 107 languages. The base model has a vocabulary of 256,000 subwords, and byte-level fallback breaks unfamiliar tokens into UTF-8 bytes. This pre-training uses the Adafactor optimizer \cite{shazeer2018adafactor} with a constant learning rate of 0.01 for the first 10,000 steps and inverse square root decay later.

\vspace*{-7pt}

\subsubsection{Byte T5 (ByT5)}

\cite{xue2022ByT5} is a token-free variation of the mT5 model which is fed UTF-8 bytes directly without any text pre-processing. This solves the limitation of handling OOV and variation in word representation due to differences in capitalization, spelling and many more. The ByT5 model uses a 256-dimensional vocabulary to directly embed the input bytes into the model's hidden size. Additionally, the model reserves three special token IDs for padding, end-of-sentence, and an unused token, which is included for convention. In this model, spans of approximately 20 bytes are masked, and the encoder is three times deeper than the decoder. Like mT5, the sequence length is set to 1024 bytes instead of tokens and is trained for 1 million steps over batches of $2^{20}$ tokens. BYT5 is robust to noise, they are beyond any language barrier for text processing and does not require any perplexing text pre-processing pipelines. Nevertheless, the byte-level approach has a disadvantage in that the byte sequences are typically longer than the original text sequences, leading to increased processing costs.

\section{Experiments\label{sec5}}

This section discusses the experiments conducted for our methodology, equipping the proposed DGT approach. We carry out experiments on various transformer models and conduct an ablation study on whether the DGT tokens cause a gain in performance. Furthermore, we conduct experiments to observe the individual performance of the models for each district's district. The experiments and ablation studies support the effectiveness of the DGT technique in transcribing Bengali regional text to IPAs.

\subsection{Experimental setup}

We compare different machine translation models to find the most efficient training approach. We explore three models: ByT5 (small version), umT5 (base version), and mT5 (base version). We determine the optimal number of training epochs for each model by monitoring their performance. We find that ByT5 required the most training at 25 epochs, while the umT5 and the mT5 achieved stable performance in 10 and 15 epochs, respectively. The training was done using the AdamW \cite{loshchilov2017decoupled} optimizer with a learning rate of $3 \times 10^{-4}$ and a weight decay of $1 \times 10^{-2}$. The loss function used is the Categorical Cross-entropy loss. The experiments were implemented in PyTorch, and the models were run on 2 NVIDIA Tesla T4 GPUs with 14.7 GB of memory each.

\subsection{Evaluation metrics}

To validate the performance of our models, we mainly use two empirical metrics for transcription tasks: the word error rate shown in Equation \ref{eq:wer} and the character error rate \ref{eq:cer}. WER provides a quantitative measure of the accuracy of the transcription system by comparing the original text with the transcribed IPA symbols. WER takes into account insertions, deletions, and substitutions of words, providing a comprehensive view of the model's performance.

\begin{equation}
    \text{Word Error Rate (WER)} = \frac{S+D+I}{S+D+W}
    \label{eq:wer}
\end{equation}
where $S$ is the number of substitutions, $D$ is the number of deletions, $I$ is the number of insertions, and $W$ is the number of correct words.

\begin{equation}
    \text{Character Error Rate (CER)} = \frac{S+D+I}{S+D+C}
    \label{eq:cer}
\end{equation}
where $C$ is the number of correct characters.

The main limitation of Word Error Rate (WER) for evaluating text-to-IPA transcription is that it only considers word-level differences without capturing the nuances of phonetic accuracy. This is addressed by using Character Error Rate (CER) as a complementary matrix that operates on a character level of the text. CER is calculated by comparing the individual characters or phonemes in the generated text with the reference text, and it provides a more granular assessment of the phonetic accuracy of the transcription.

\vspace*{-7pt}

\subsection{Results}

Table \ref{tab1} illustrates the performance of our models with the District Guided Tokens. The results are recorded for the test set. We can see that the ByT5 model performs the best with a WER of 1.2\% and a CER of 0.4\%. This means that in a generation of 100 IPA words, roughly 1.2 words and 0.4 characters will result in an error. It demonstrates the ability of the ByT5 model to excel in its generation task even with a significant presence of OOV words. The umT5-base also performs relatively well in the presence of OOV words and competes in performance with the ByT5 model by having a WER of 2.39\% and CER of 1.01\%. We will see in the ablation studies a few examples where the umT5 excelled slightly over the ByT5 model. However, its predecessor architecture, the mT5, performs at a WER of 27.79\% and 5.63\%. This is due to the umT5's unique relative positioning bias ability over the mT5's shared bias ability. This ability gives umT5 the advantage of learning unique values at each layer in the model, giving it a better understanding of the Bengali language and the positional patterns of its words.

\begin{table}[t]
\def\arraystretch{1.5}%
\setlength\tabcolsep{1.4em}
\centering
\caption{The word error rate and character error rate, in percentage, for the DGT technique.}\label{tab1}
\begin{tabular}{|c|c|c|}
\hline
\textbf{Model} &  \textbf{Word Error Rate (\%)}  & \textbf{Character Error Rate (\%)}\\ \hline\

\textbf{byt5-small} & \textbf{1.24}  & \textbf{0.42} \\ \hline
\textbf{umT5-base} & 2.39 & 1.01 \\ \hline
\textbf{mT5-base} & 27.79 & 5.63 \\ \hline

\end{tabular}
\end{table}

\begin{table}[t]
\def\arraystretch{1.5}%
\setlength\tabcolsep{1.6em}
\centering
\caption{The word error rate and character error rate, in percentage, for the DGT and without DGT technique.}\label{dgtnodgt}
\begin{tabular}{|c|cc|cc|}
\hline
\multirow{2}{*}{\textbf{Model}} & \multicolumn{2}{c|}{\textbf{With DGT}} & \multicolumn{2}{c|}{\textbf{Without DGT}} \\ \cline{2-5} 
& \multicolumn{1}{c|}{WER (\%)}     & CER (\%)     & \multicolumn{1}{c|}{WER (\%)}      & CER (\%)       \\ \hline
\textbf{byt5-small}                   & \multicolumn{1}{c|}{1.24}  & 0.42  & \multicolumn{1}{c|}{1.80}    & 0.94    \\ \hline
\textbf{umT5-base}              & \multicolumn{1}{c|}{2.39}  & 1.01  & \multicolumn{1}{c|}{3.00}     & 1.54    \\ \hline
\textbf{mT5-base}               & \multicolumn{1}{c|}{27.79}  & 5.63  & \multicolumn{1}{c|}{28.25}         &    5.94       \\ \hline

\end{tabular}
\end{table}

\vspace*{-\baselineskip}

\begin{table}[t]
\def\arraystretch{1.5}%
\setlength\tabcolsep{0.45em}
\centering
\caption{The word error rate and character error rate, in percentage, of the 6 districts with the DGT technique.}\label{diswithdgt}
\begin{tabular}{|c|cc|cc|cc|}
\hline
\multirow{2}{*}{\textbf{District}} & \multicolumn{2}{c|}{\textbf{byt5-small}} & \multicolumn{2}{c|}{\textbf{umT5-base}} & \multicolumn{2}{c|}{\textbf{mT5-base}} \\ \cline{2-7} 
& \multicolumn{1}{c|}{WER (\%)}      & CER (\%)     & \multicolumn{1}{c|}{WER (\%)}      & CER (\%)    & \multicolumn{1}{c|}{WER (\%)}     & CER (\%)     \\ \hline
\textbf{Rangpur}                   & \multicolumn{1}{c|}{0.045}  & 0.021  & \multicolumn{1}{c|}{0.765}  & 0.421 & \multicolumn{1}{c|}{22.43}       & 4.348       \\ \hline
\textbf{Narail}                    & \multicolumn{1}{c|}{0.032}  & 0.018  & \multicolumn{1}{c|}{0.588}  & 0.394 & \multicolumn{1}{c|}{26.22}       & 4.919       \\ \hline
\textbf{Kishoreganj}               & \multicolumn{1}{c|}{0.584}  & 0.701  & \multicolumn{1}{c|}{1.304}  & 0.952 & \multicolumn{1}{c|}{26.91}       & 5.420       \\ \hline
\textbf{Narsingdi}                 & \multicolumn{1}{c|}{0.097}  & 0.041  & \multicolumn{1}{c|}{0.685}  & 0.225 & \multicolumn{1}{c|}{26.67}       & 4.572       \\ \hline
\textbf{Tangail}                   & \multicolumn{1}{c|}{0.076}  & 0.031  & \multicolumn{1}{c|}{0.796}  & 0.467 & \multicolumn{1}{c|}{24.97}       & 4.732       \\ \hline
\textbf{Chittagong}                & \multicolumn{1}{c|}{1.106}  & 0.85   & \multicolumn{1}{c|}{2.068}  & 1.237 & \multicolumn{1}{c|}{30.49}       & 6.293       \\ \hline
\end{tabular}
\end{table}

\vspace*{8pt}

\subsection{Ablation study}

\subsubsection{Performance of the District Guided Tokens}

We investigated the effectiveness between DGT and non-DGT models. To ensure a fair comparison, both methods were trained for the same to even more epochs than the DGT models. Interestingly, even with extensive training, the non-DGT models never surpassed the performance of the DGT models. Therefore, we recorded the best performances of the non-DGT models within the same number of epochs trained. The results are illustrated in Table \ref{dgtnodgt}.

\vspace*{-\baselineskip}

\subsubsection{District-wise Performance}

We evaluated the DGT models on individual districts within the test set. This is illustrated in Table \ref{diswithdgt}. The Narail district emerged as the easiest for the models to generate the IPAs, achieving the lowest WER and CER scores across all models. The Chittagong district proved the most challenging, consistently exhibiting the highest WER and CER scores. This difficulty likely stems from pronunciation and spelling variations of common words and characters in the Chittagong district compared to other districts. 


\section{Discussion\label{sec6}}


\subsection{Performance}

We have recorded the outputs for the ByT5 and the umT5 models on selective examples illustrated in Appendices \ref{appendix:casesbyt5} and \ref{appendix:casesumt5}. Both models make errors on the same words. The ByT5 generates the right IPA for most cases than the umT5 as shown by the examples of the Rangpur district. However, the ByT5 model skips certain words, represented in blue, for the Narsingdi example, but the umT5 catches these words. Moreover, the ByT5 generates one erroneous character in the Chittagong example, which the umT5 does not. We mitigated this problem by applying a word-by-word inference for the ByT5 model. First, we tokenized the sentence without adding the DGT at the start. Next, we add the DGT at the beginning of each word during the word-by-word inference. This mitigates the error of the missed IPAs by ByT5 during generation. However, compared to the performance of the original DGT-based ByT5 inference, it lacks in performance by 0.1\% achieving a WER of 1.3\% and a CER of 0.52\%.

\subsection{Limitations}

While this method boosts the performance of regional IPA transcription, the DGT is programmatically added into the input text that is to be transcribed. Future work would address the dynamic selection of the DGT through a lightweight MLP. Moreover, the quantity of the current finetuned models, and the experiments that surround those models, are early Language models and not Large Language Models (LLMs). Future work would address the capabilities of LLMs in regional Bengali transcription.

\section{Conclusion\label{sec7}}

In this study, we have introduced the District Guided Tokens (DGT) for transcribing Bengali regional districts of six districts to IPA. We have proven that the utilization of DGT has significantly improved performance across all three multilingual transformer models: umT5, mT5, and ByT5, compared to not using it. The DGT performs noticeably better with the token-free ByT5 transformer model, as it can adequately handle OOV and variation in word representations, which is instrumental in learning phonetic diversity among different districts. Our study shows that equipping DGT in multilingual transformers could remarkably boost performance in IPA transcription of the Bengali language with multiple districts, opening the window to investigate local parlance. We have released all code and pre-trained models used in this paper to facilitate future work and accelerate the incorporation of the Bengali language with regional districts into mainstream artificial intelligence applications.

\bibliographystyle{splncs04}
\bibliography{samplepaper}

\begin{thebibliography}{10}
\providecommand{\url}[1]{\texttt{#1}}
\providecommand{\urlprefix}{URL }
\providecommand{\doi}[1]{https://doi.org/#1}

\bibitem{bhattacharjee-etal-2023-banglanlg}
Bhattacharjee, A., Hasan, T., Ahmad, W.U., Shahriyar, R.: {B}angla{NLG} and {B}angla{T}5: Benchmarks and resources for evaluating low-resource natural language generation in {B}angla. In: Vlachos, A., Augenstein, I. (eds.) Findings of the Association for Computational Linguistics: EACL 2023. pp. 726--735. Association for Computational Linguistics, Dubrovnik, Croatia (May 2023). \doi{10.18653/v1/2023.findings-eacl.54}, \url{https://aclanthology.org/2023.findings-eacl.54}

\bibitem{chung2022unimax}
Chung, H.W., Garcia, X., Roberts, A., Tay, Y., Firat, O., Narang, S., Constant, N.: Unimax: Fairer and more effective language sampling for large-scale multilingual pretraining. In: The Eleventh International Conference on Learning Representations (2022)

\bibitem{dave2021neural}
Dave, S., Singh, A.K., AP, D.P., Lall, P.B.: Neural compound-word (sandhi) generation and splitting in sanskrit language. In: Proceedings of the 3rd ACM India Joint International Conference on Data Science \& Management of Data (8th ACM IKDD CODS \& 26th COMAD). pp. 171--177 (2021)

\bibitem{devlin2018bert}
Devlin, J., Chang, M.W., Lee, K., Toutanova, K.: Bert: Pre-training of deep bidirectional transformers for language understanding. arXiv preprint arXiv:1810.04805  (2018)

\bibitem{faria2023vashantor}
Faria, F.T.J., Moin, M.B., Wase, A.A., Ahmmed, M., Sani, M.R., Muhammad, T.: Vashantor: A large-scale multilingual benchmark dataset for automated translation of bangla regional dialects to bangla language. arXiv preprint arXiv:2311.11142  (2023)

\bibitem{fatema2024ipa}
Fatema, K., Haider, F.D., Turpa, N.F., Azmal, T., Ahmed, S., Hasan, N., Rahman, M.A., Sarkar, B.K., Jahin, A., Hassan, M.R., Zihad, M.F., Faruque, R.S., Sushmit, A., Imtiaz, M., Sadeque, F., Rahman, S.S.: Ipa transcription of bengali texts (2024)

\bibitem{loshchilov2017decoupled}
Loshchilov, I., Hutter, F.: Decoupled weight decay regularization. arXiv preprint arXiv:1711.05101  (2017)

\bibitem{mukherjeebengali}
Mukherjee, S., Mandal, S.K.D.: A bengali hmm based speech synthesis system. In: International Conference on Speech Database and Assessments (Oriental COCOSDA). pp. 255--259 (2012)

\bibitem{2019t5}
Raffel, C., Shazeer, N., Roberts, A., Lee, K., Narang, S., Matena, M., Zhou, Y., Li, W., Liu, P.J.: Exploring the limits of transfer learning with a unified text-to-text transformer. arXiv e-prints  (2019)

\bibitem{roberts2020knowledge}
Roberts, A., Raffel, C., Shazeer, N.: How much knowledge can you pack into the parameters of a language model? (2020)

\bibitem{sen2022bangla}
Sen, O., Fuad, M., Islam, M.N., Rabbi, J., Masud, M., Hasan, M.K., Awal, M.A., Fime, A.A., Fuad, M.T.H., Sikder, D., et~al.: Bangla natural language processing: A comprehensive analysis of classical, machine learning, and deep learning-based methods. IEEE Access  \textbf{10},  38999--39044 (2022)

\bibitem{shazeer2020glu}
Shazeer, N.: Glu variants improve transformer. arXiv preprint arXiv:2002.05202  (2020)

\bibitem{shazeer2018adafactor}
Shazeer, N., Stern, M.: Adafactor: Adaptive learning rates with sublinear memory cost (2018)

\bibitem{sutskever2014sequence}
Sutskever, I., Vinyals, O., Le, Q.V.: Sequence to sequence learning with neural networks. Advances in neural information processing systems  \textbf{27} (2014)

\bibitem{taguchi2023universal}
Taguchi, C., Sakai, Y., Haghani, P., Chiang, D.: Universal automatic phonetic transcription into the international phonetic alphabet. arXiv preprint arXiv:2308.03917  (2023)

\bibitem{taguchi2023}
Taguchi, C., Sakai, Y., Haghani, P., Chiang, D.: Universal automatic phonetic transcription into the international phonetic alphabet. arXiv preprint arXiv:2308.03917  (2023)

\bibitem{xue2022ByT5}
Xue, L., Barua, A., Constant, N., Al-Rfou, R., Narang, S., Kale, M., Roberts, A., Raffel, C.: Byt5: Towards a token-free future with pre-trained byte-to-byte models. Transactions of the Association for Computational Linguistics  \textbf{10},  291--306 (2022)

\bibitem{xue-etal-2021-mt5}
Xue, L., Constant, N., Roberts, A., Kale, M., Al-Rfou, R., Siddhant, A., Barua, A., Raffel, C.: m{T}5: A massively multilingual pre-trained text-to-text transformer. In: Toutanova, K., Rumshisky, A., Zettlemoyer, L., Hakkani-Tur, D., Beltagy, I., Bethard, S., Cotterell, R., Chakraborty, T., Zhou, Y. (eds.) Proceedings of the 2021 Conference of the North American Chapter of the Association for Computational Linguistics: Human Language Technologies. pp. 483--498. Association for Computational Linguistics, Online (Jun 2021). \doi{10.18653/v1/2021.naacl-main.41}, \url{https://aclanthology.org/2021.naacl-main.41}

\end{thebibliography}

\newpage

\appendix
\section{Appendices}

\subsection{Longest sentence length example\label{appendix:longestsen}}
\vspace*{-\baselineskip}
\vspace*{-\baselineskip}
\begin{table}[h]
\centering
\begin{tabular}{|p{\textwidth}|}
\hline
\begin{tabular}[p{\textwidth}]{@{}p{\textwidth}@{}}\textbf{District:} Rangpur\\ \\ \textbf{Bengali Text:} \bn{ব্যাবহার যার ভালো থাকবে তার জাগা জমিন এর দরকার হয় না। অইযে এই যাগা টা বেঁচে খাইসে, বেচে আরো উপরের যায়গা খানত গেলো। রংপুর এর যাগা খান বলে বেঁচার চিন্তা ভাবনা কইরবার লাগসে ওখান বেঁচে বলে কোটে বলে টাইস ওলা বাড়ি বলে দিবে। অইল্লা করিয়া কোনো লাভ নাই, অইল্লা করিয়া কোনো লাভ নাই। ট্রাক আসিল না ট্রাকও বেচাইসে।}\\ \\ \textbf{IPA:} \ipa{bɛbohɐɾ ɟɐɾ bʱɐlo t̪ʰɐkbe t̪ɐɾ ɟɐgɐ ɟomɪn eɾ d̪ɔɾkɐɾ hɔe̯ nɐ| ɔ͡ɪ̯ɟe e͡ɪ̯ ɟɐgɐ tɐ bẽce kʰɐ͡ɪ̯ʃe, bece ɐɾo upoɾeɾ ɟɐʲgɐ kʰɐnɔt̪o gelo| ɾɔŋpuɾ eɾ ɟɐgɐ kʰɐn bole bẽcɐɾ cɪnt̪ɐ bʱɐbnɐ ko͡ɪ̯ɾbɐɾ lɐgʃe okʰɐn bẽce bole kote bole tɐ͡ɪ̯ʃ olɐ bɐɽɪ bole d̪ɪbe। ɔ͡ɪ̯llɐ koɾɪʲɐ kono lɐbʱ nɐ͡ɪ̯, ɔ͡ɪ̯llɐ koɾɪʲɐ kono lɐbʱ nɐ͡ɪ̯| tɾɐk ɐʃɪl nɐ tɾɐko becɐ͡ɪ̯ʃe}\bn{।}\end{tabular} \\ \hline
\end{tabular}
\end{table}

\vspace*{-\baselineskip}
\vspace*{-\baselineskip}

\subsection{Longest IPA length example\label{appendix:longestipa}}
\vspace*{-\baselineskip}
\vspace*{-\baselineskip}
\begin{table}[h]
\centering
\begin{tabular}{|p{\textwidth}|}
\hline
\begin{tabular}[p{\textwidth}]{@{}p{\textwidth}@{}}\textbf{District:} Chittagong\\ \\ \textbf{Bengali Text:} \bn{হেইয়ি, হেইয়েনে আবার বুমি গরি দিয়ি, আবার গুম গিয়িগই, তারফর বেইননা চাইরগান নে ছওয়ান কেতা জমা অইয়ে, ইন লইয়োনে নানু আর আম্মু ধুতো গেইয়ে গই, আই উড়ি, আবার গুম গিয়িগই, আত পা শক্ত অই গিয়িগই, আই আর কিছু গরিত ন-ফারি, এন্ডে ইবা অইলদে মুনতাকিম, ইবারতুন ও অসুক অই গিয়ে, ইতেও গুম যার উড়ের,  আরতো আরা কিছু গরিত ন-ফারি।}\\ \\ \textbf{IPA:} \ipa{he͡ɪ̯ʲɪ, he͡ɪ̯ʲene ɐbɐɾ bumɪ goɾɪ d̪ɪʲɪ, ɐbɐɾ gum gɪʲɪgɪ, t̪ɐɾpʰɔɾ be͡ɪ̯nnɐ cɐ͡ɪ̯ɾgɐn ne cʰɔ͡o̯ʲɐn ket̪ɐ ɟɔmɐ ɔ͡ɪ̯ʲe, ɪn lɔ͡ɪ̯ʲone nɐnu ɐɾ ɐmmu d̪ʱut̪o ge͡ɪ̯ʲe go͡ɪ̯, ɐ͡ɪ̯ uɽɪ, ɐbɐɾ gum gɪʲɪgɪ, ɐt̪o pɐ ʃɔkt̪o ɔ͡ɪ̯ gɪʲɪgɪ, ɐ͡ɪ̯ ɐɾ kɪcʰu goɾɪt̪o-pʰɐɾɪ, ɛnde ɪbɐ ɔ͡ɪ̯ld̪e munot̪ɐkɪm, ɪbɐɾt̪un o ɔʃuk ɔ͡ɪ̯ gɪʲe, ɪt̪e͡o̯ gum ɟɐɾ uɽeɾ,  ɐɾɔt̪o ɐɾɐ kɪcʰu goɾɪt̪o-pʰɐɾɪ}\bn{।}\end{tabular} \\ \hline
\end{tabular}
\end{table}

\clearpage

\subsection{Selective cases of the ByT5 with the DGT technique. Red colours are erroneous words, blue colours are missed words.\label{appendix:casesbyt5}}
\vspace*{-\baselineskip}
\begin{table}[h]
\centering
\def\arraystretch{1.3}%
\setlength\tabcolsep{0.45em}
\begin{tabular}{|l|p{0.85\textwidth}|}
\hline
\textbf{District}             & \textbf{Examples}                                                                                                                                                                                                                                                                                                                                      \\ \hline
\multicolumn{1}{|c|}{Rangpur} & \begin{tabular}[p{0.85\textwidth}]{@{}p{0.85\textwidth}@{}}\textbf{Text:} \bn{খরচ করিম এই যে ওজার খরচটা করিম, এই খরচটা করার জন্যে এক হাজার টেহা খুঁজবের নাগছোং।}\\ \textbf{Pred:} \ipa{kʰɔɾoc koɾɪm e͡ɪ̯ ɟe oɟɐɾ kʰɔɾoctɐ koɾɪm, e͡ɪ̯ kʰɔɾoctɐ kɔɾɐɾ ɟonne ɛk hɐɟɐɾ tehɐ \textcolor{red}{kʰũɟbeɾ} nɐgcʰoŋ}\bn{।}\\ \textbf{GT:} \ipa{kʰɔɾoc koɾɪm e͡ɪ̯ ɟe oɟɐɾ kʰɔɾoctɐ koɾɪm, e͡ɪ̯ kʰɔɾoctɐ kɔɾɐɾ ɟonne ɛk hɐɟɐɾ tehɐ \textcolor{red}{kʰũɟɔbeɾ} nɐgcʰoŋ}\bn{।}\end{tabular} \\ \hline

\multicolumn{1}{|c|}{Narail} & \begin{tabular}[p{0.85\textwidth}]{@{}p{0.85\textwidth}@{}}\textbf{Text:} \bn{না, আমি দৌড়ইছি জুতো পইড়ে তা আমার পার তলে আগুন ধইরে যাওয়ার মত অবস্থা।}\\ \textbf{Pred:} \ipa{nɐ, ɐmɪ \textcolor{red}{d̪o͡u̯ɽo͡ɪ̯cʰɪ} ɟut̪o po͡ɪ̯ɽe t̪ɐ ɐmɐɾ pɐɾ t̪ɔle ɐgun d̪ʱo͡ɪ̯ɾe ɟɐ͡o̯ʷɐɾ mɔt̪ ɔbost̪ʰɐ}\bn{।}\\ \textbf{GT:} \ipa{nɐ, ɐmɪ \textcolor{red}{d̪o͡u̯no͡ɪ̯cʰɪ} ɟut̪o po͡ɪ̯ɽe t̪ɐ ɐmɐɾ pɐɾ t̪ɔle ɐgun d̪ʱo͡ɪ̯ɾe ɟɐ͡o̯ʷɐɾ mɔt̪ ɔbost̪ʰɐ}\bn{।}\end{tabular} \\ \hline

\multicolumn{1}{|c|}{Kishoreganj} & \begin{tabular}[p{0.85\textwidth}]{@{}p{0.85\textwidth}@{}}\textbf{Text:} \bn{এইযে র‍্যাপটা, এইযে ইয়ও কয়েকদিন আগে বাও করছিল না ডাহার?}\\ \textbf{Pred:} \ipa{e͡ɪ̯ɟe \textcolor{red}{ɾɛptɐ}, e͡ɪ̯ɟe ɪʲo kɔʲekd̪ɪn ɐge bɐ͡o̯ koɾcʰɪlo nɐ dɐhɐɾ?}\\ \textbf{GT:} \ipa{e͡ɪ̯ɟe \textcolor{red}{ɾɛpɔtɐ}, e͡ɪ̯ɟe ɪʲo kɔʲekd̪ɪn ɐge bɐ͡o̯ koɾcʰɪlo nɐ dɐhɐɾ?} \end{tabular} \\ \hline

\multicolumn{1}{|c|}{Narsingdi} & \begin{tabular}[p{0.85\textwidth}]{@{}p{0.85\textwidth}@{}}\textbf{Text:} \bn{একটু ফরে-ফরে যাইতে থায়ে, আইতে থায়ে, আইতেই থায়ে, বিদ্যুৎ অফিসের লোকেরা ফাইসেও, যেমুন ধরেন আনুপাতিক হারে কারেন যায় বলে।}\\ \textbf{Pred:} \ipa{ektu pʰɔɾe-\textcolor{red}{pʰɔɾe} ɟɐ͡ɪ̯t̪e t̪ʰɐʲe, ɐ͡ɪ̯t̪e͡ɪ̯ t̪ʰɐʲe, bɪd̪d̪ut̪ ɔpʰɪseɾ lokeɾɐ pʰɐ͡ɪ̯se͡o̯, ɟemun d̪ʱɔɾen ɐnupɐt̪ɪk hɐɾe kɐɾen ɟɐe̯ bole}\bn{।}\\ \textbf{GT:} \ipa{ektu pʰɔɾe-\textcolor{red}{pʰɾe} ɟɐ͡ɪ̯t̪e t̪ʰɐʲe, \textcolor{blue}{ɐ͡ɪ̯t̪e t̪ʰɐʲe,} ɐ͡ɪ̯t̪e͡ɪ̯ t̪ʰɐʲe, bɪd̪d̪ut̪ ɔpʰɪseɾ lokeɾɐ pʰɐ͡ɪ̯se͡o̯, ɟemun d̪ʱɔɾen ɐnupɐt̪ɪk hɐɾe kɐɾen ɟɐe̯ bole}\bn{।}\end{tabular} \\ \hline

\multicolumn{1}{|c|}{Tangail} & \begin{tabular}[p{0.85\textwidth}]{@{}p{0.85\textwidth}@{}}\textbf{Text:} \bn{আপনের মানে কড়া থাকোন নাগবো যদি আপনেও যদি ভালো কইরা না দেইন তাইলে যা অনেক অনেক মুরুক্ষতার কারণে মানুষ ই অয়।}\\ \textbf{Pred:} \ipa{ɐponeɾ mɐne kɔɽɐ t̪ʰɐkon nɐgobo ɟod̪ɪ \textcolor{red}{ɐpne͡o̯} ɟod̪ɪ bʱɐlo ko͡ɪ̯ɾɐ nɐ d̪e͡ɪ̯n t̪ɐ͡ɪ̯le ɟɐ ɔnek ɔnek muɾukkʰot̪ɐɾ kɐɾone mɐnuʃ ɪ ɔe̯}\bn{।}\\ \textbf{GT:} \ipa{ɐponeɾ mɐne kɔɽɐ t̪ʰɐkon nɐgobo ɟod̪ɪ \textcolor{red}{ɐpone͡o̯} ɟod̪ɪ bʱɐlo ko͡ɪ̯ɾɐ nɐ d̪e͡ɪ̯n t̪ɐ͡ɪ̯le ɟɐ ɔnek ɔnek muɾukkʰot̪ɐɾ kɐɾone mɐnuʃ ɪ ɔe̯}\bn{।}\end{tabular} \\ \hline

\multicolumn{1}{|c|}{Chittagong} & \begin{tabular}[p{0.85\textwidth}]{@{}p{0.85\textwidth}@{}}\textbf{Text:} \bn{আঁই তো ওর তুন টেঁয়ার খতা ফুছ গরি, ইতে ফাঁচশো টেঁয়া খইবু আঁরে।}\\ \textbf{Pred:} \ipa{ɐ̃ɪ t̪o oɾ t̪un tẽʲɐɾ kʰɔt̪ɐ \textcolor{red}{pʰucʰo} goɾɪ, ɪt̪e pʰɐ̃cʃo tẽʲɐ kʰo͡ɪ̯bu ɐ̃ɾe}\bn{।}\\ \textbf{GT:} \ipa{ɐ̃ɪ t̪o oɾ t̪un tẽʲɐɾ kʰɔt̪ɐ \textcolor{red}{pʰucʰ} goɾɪ, ɪt̪e pʰɐ̃cʃo tẽʲɐ kʰo͡ɪ̯bu ɐ̃ɾe}\bn{।}\end{tabular} \\ \hline

\end{tabular}
\end{table}

\clearpage

\subsection{Selective cases of the umT5 with the DGT technique. Red colours are erroneous words.\label{appendix:casesumt5}}
\vspace*{-\baselineskip}
\begin{table}[h]
\centering
\def\arraystretch{1.3}%
\setlength\tabcolsep{0.45em}
\begin{tabular}{|l|p{0.85\textwidth}|}
\hline
\textbf{District}             & \textbf{Examples}                                                                                                                                                                                                                                                                                     \\ \hline
\multicolumn{1}{|c|}{Rangpur} & \begin{tabular}[p{0.85\textwidth}]{@{}p{0.85\textwidth}@{}}\textbf{Text:} \bn{খরচ করিম এই যে ওজার খরচটা করিম, এই খরচটা করার জন্যে এক হাজার টেহা খুঁজবের নাগছোং।}\\ \textbf{Pred:} \ipa{kʰɔɾoc koɾɪm e͡ɪ̯ ɟe oɟɐɾ \textcolor{red}{kʰɔɾctɐ} koɾɪm, e͡ɪ̯ \textcolor{red}{kʰɔɾctɐ} kɔɾɐɾ ɟonne ɛk hɐɟɐɾ tehɐ kʰũɟɔbeɾ nɐgcʰoŋ}\bn{।}
\\ \textbf{GT:} \ipa{kʰɔɾoc koɾɪm e͡ɪ̯ ɟe oɟɐɾ \textcolor{red}{kʰɔɾoctɐ} koɾɪm, e͡ɪ̯ \textcolor{red}{kʰɔɾoctɐ} kɔɾɐɾ ɟonne ɛk hɐɟɐɾ tehɐ kʰũɟɔbeɾ nɐgcʰoŋ}\bn{।}\end{tabular} \\ \hline

\multicolumn{1}{|c|}{Narail} & \begin{tabular}[p{0.85\textwidth}]{@{}p{0.85\textwidth}@{}}\textbf{Text:} \bn{না, আমি দৌড়ইছি জুতো পইড়ে তা আমার পার তলে আগুন ধইরে যাওয়ার মত অবস্থা।}\\ \textbf{Pred:} \ipa{nɐ, ɐmɪ \textcolor{red}{d̪o͡u̯ɽɪcʰɪ} ɟut̪o po͡ɪ̯ɽe t̪ɐ ɐmɐɾ pɐɾ t̪ɔle ɐgun d̪ʱo͡ɪ̯ɾe ɟɐ͡o̯ʷɐɾ mɔt̪ ɔbost̪ʰɐ}\bn{।}\\ \textbf{GT:} \ipa{nɐ, ɐmɪ \textcolor{red}{d̪o͡u̯no͡ɪ̯cʰɪ} ɟut̪o po͡ɪ̯ɽe t̪ɐ ɐmɐɾ pɐɾ t̪ɔle ɐgun d̪ʱo͡ɪ̯ɾe ɟɐ͡o̯ʷɐɾ mɔt̪ ɔbost̪ʰɐ}\bn{।}\end{tabular} \\ \hline

\multicolumn{1}{|c|}{Kishoreganj} & \begin{tabular}[p{0.85\textwidth}]{@{}p{0.85\textwidth}@{}}\textbf{Text:} \bn{এইযে র‍্যাপটা, এইযে ইয়ও কয়েকদিন আগে বাও করছিল না ডাহার?}\\ \textbf{Pred:} \ipa{e͡ɪ̯ɟe \textcolor{red}{ɾɛptɐ}, e͡ɪ̯ɟe ɪʲo kɔʲekd̪ɪn ɐge bɐ͡o̯ koɾcʰɪlo nɐ dɐhɐɾ?}\\ \textbf{GT:} \ipa{e͡ɪ̯ɟe \textcolor{red}{ɾɛpɔtɐ}, e͡ɪ̯ɟe ɪʲo kɔʲekd̪ɪn ɐge bɐ͡o̯ koɾcʰɪlo nɐ dɐhɐɾ?} \end{tabular} \\ \hline

\multicolumn{1}{|c|}{Narsingdi} & \begin{tabular}[p{0.85\textwidth}]{@{}p{0.85\textwidth}@{}}\textbf{Text:} \bn{একটু ফরে-ফরে যাইতে থায়ে, আইতে থায়ে, আইতেই থায়ে, বিদ্যুৎ অফিসের লোকেরা ফাইসেও, যেমুন ধরেন আনুপাতিক হারে কারেন যায় বলে।}\\ \textbf{Pred:} \ipa{ektu pʰɔɾe-\textcolor{red}{pʰɔɾe} ɟɐ͡ɪ̯t̪e t̪ʰɐʲe, ɐ͡ɪ̯t̪e t̪ʰɐʲe, ɐ͡ɪ̯t̪e͡ɪ̯ t̪ʰɐʲe, bɪd̪d̪ut̪ ɔpʰɪseɾ lokeɾɐ pʰɐ͡ɪ̯ʃe͡o̯, ɟemun d̪ʱɔɾen ɐkupɐt̪ɪk hɐɾe kɐɾen ɟɐe̯ bole}\bn{।}\\ \textbf{GT:} \ipa{ektu pʰɔɾe-\textcolor{red}{pʰɾe} ɟɐ͡ɪ̯t̪e t̪ʰɐʲe, ɐ͡ɪ̯t̪e t̪ʰɐʲe, ɐ͡ɪ̯t̪e͡ɪ̯ t̪ʰɐʲe, bɪd̪d̪ut̪ ɔpʰɪseɾ lokeɾɐ pʰɐ͡ɪ̯se͡o̯, ɟemun d̪ʱɔɾen ɐnupɐt̪ɪk hɐɾe kɐɾen ɟɐe̯ bole}\bn{।}\end{tabular} \\ \hline

\multicolumn{1}{|c|}{Tangail} & \begin{tabular}[p{0.85\textwidth}]{@{}p{0.85\textwidth}@{}}\textbf{Text:} \bn{আপনের মানে কড়া থাকোন নাগবো যদি আপনেও যদি ভালো কইরা না দেইন তাইলে যা অনেক অনেক মুরুক্ষতার কারণে মানুষ ই অয়।}\\ \textbf{Pred:} \ipa{ɐponeɾ mɐne kɔɽɐ t̪ʰɐkon nɐgobo ɟod̪ɪ \textcolor{red}{ɐpne͡o̯} ɟod̪ɪ bʱɐlo ko͡ɪ̯ɾɐ nɐ d̪e͡ɪ̯n t̪ɐ͡ɪ̯le ɟɐ ɔnek ɔnek muɾukkʰot̪ɐɾ kɐɾone mɐnuʃ ɪ ɔe̯}\bn{।}\\ \textbf{GT:} \ipa{ɐponeɾ mɐne kɔɽɐ t̪ʰɐkon nɐgobo ɟod̪ɪ \textcolor{red}{ɐpone͡o̯} ɟod̪ɪ bʱɐlo ko͡ɪ̯ɾɐ nɐ d̪e͡ɪ̯n t̪ɐ͡ɪ̯le ɟɐ ɔnek ɔnek muɾukkʰot̪ɐɾ kɐɾone mɐnuʃ ɪ ɔe̯}\bn{।}\end{tabular} \\ \hline

\multicolumn{1}{|c|}{Chittagong} & \begin{tabular}[p{0.85\textwidth}]{@{}p{0.85\textwidth}@{}}\textbf{Text:} \bn{আঁই তো ওর তুন টেঁয়ার খতা ফুছ গরি, ইতে ফাঁচশো টেঁয়া খইবু আঁরে।}\\ \textbf{Pred:} \ipa{ɐ̃ɪ t̪o oɾ t̪un tẽʲɐɾ kʰɔt̪ɐ pʰucʰ goɾɪ, ɪt̪e pʰɐ̃cʃo tẽʲɐ kʰo͡ɪ̯bu ɐ̃ɾe}\bn{।}\\ \textbf{GT:} \ipa{ɐ̃ɪ t̪o oɾ t̪un tẽʲɐɾ kʰɔt̪ɐ pʰucʰ goɾɪ, ɪt̪e pʰɐ̃cʃo tẽʲɐ kʰo͡ɪ̯bu ɐ̃ɾe}\bn{।}\end{tabular} \\ \hline

\end{tabular}
\end{table}

\end{document}